# An Experiment in Morphological Development for Learning ANN Based Controllers

M. Naya-Varela
*Integrated Group for Engineering Research*
*CITIC (Centre for Information and*
*Communications Technology*
*Research)Universidade da Coruña*
A Coruña, Spain.
martin.naya@udc.es

A. Faina.
*Robotics, Evolution and Art Lab (REAL)*
*Computer Science Department*
*IT University of Copenhagen*
Copenhagen, Denmark
anfv@itu.dk

R. J. Duro
*Integrated Group for Engineering Research*
*CITIC (Centre for Information and*
*Communications Technology*
*Research)Universidade da Coruña*
A Coruña, Spain.
richard@udc.es

*Abstract*— Morphological development is part of the way any human or animal learns. The learning processes starts with the morphology at birth and progresses through changing morphologies until adulthood is reached. Biologically, this seems to facilitate learning and make it more robust. However, when this approach is transferred to robotic systems, the results found in the literature are inconsistent: morphological development does not provide a learning advantage in every case. In fact, it can lead to poorer results than when learning with a fixed morphology. In this paper we analyze some of the issues involved by means of a simple, but very informative experiment in quadruped walking. From the results obtained an initial series of insights on when and under what conditions to apply morphological development for learning are presented.

I. INTRODUCTION.

While control architectures and other information processing approaches are certainly important for robotic intelligence, our understanding of how this intelligence comes about has expanded in the last decades to include the morphology of the robot and its environment, as well as their mutual interactions [1]–[3], The field of Artificial Embodied Intelligence (AEI) [4], [5], which postulates robot intelligence as the result of the interaction between brain, morphology and the environment the robot must operate in, is growing. Currently, this view on the emergence of intelligence has expanded and now it tries to address the fact that intelligent systems must be able to operate in sequences of environments that are generally unknown at design time [6]. In other words, we are facing open-ended learning problems and, by definition, these types of problems imply that robots cannot be completely defined at design time, as, at that time, we do not know what skills the robot will require in order to achieve its purpose. In fact, not even the goals that need achieving are known.

Developmental Robotics (DR) [7] is one of the approaches proposed to try to address these issues. It is based on the idea that robots can autonomously acquire an increasingly complex set of sensorimotor and mental capabilities through the interaction between its body and brain in the sequence of domains it is exposed to during its lifetime. Cangelosi defines this field as [8]:

"Developmental robotics is the interdisciplinary approach to the autonomous design of behavioral and cognitive capabilities in artificial agents that takes direct inspiration from the developmental principles and mechanisms observed in the natural cognitive systems of children "

The field focuses on the development of experience and motor skills inspired by those of humans, that emerge and grow from childhood to adulthood. A very good survey has been published by Asada et al. [7] focusing on infant development of higher cognitive functions, such as empathy or imitation. The paper provides a description of how these functions emerge in humans and how they have been implemented in a robotic system, assuming a fixed morphology for the robot during its lifetime.

On the other hand, lifelong learning in robotics is concerned with how to improve efficiency in open-ended learning through the reuse and adaptation of previously learnt knowledge. Thrun and Mitchel [9] postulated that lifelong learning should provide for faster and more robust learning than simply handling each learning process independently. This is obviously very compatible with the concept of Developmental Robotics. However most of the research that has been carried out in this field has focused on Cognitive Developmental Robotics (CDR) [10]. It has concentrated on the development of cognition [7], [11], within a fixed body and has mostly ignored the development of the body of the system as a complementary tool.

In other words, there is a missing piece in this whole approach and that is the fact that in most living beings their bodies also undergo modifications during their lifetime. That is, it is not only the cognitive part of the system that changes during development but also the body and these modifications seem to provide an advantage to the individual for learning in complex open-ended settings.

In biological systems, the importance of morphology arises from the interaction between cognition, the body, the environment, and how this interaction can be harnessed to improve the survival chances of the individual. Like biological systems, robots are physical entities operating in the real world. They have innate properties, such as mass, length, stiffness, etc., and they suffer the consequences of them: they are affected by friction, gravity, inertia, etc. Traditionally the effect of the body on cognition was deemed to be irrelevant. However, recent studies have demonstrated that the body and environmental interaction are necessary to develop communicative and cognitive skills, among others [12], [13]. In fact, as indicated by Lakoff and Núñez [14], a body is even required for activities that we take as very abstract, such as mathematical thinking.

Pfeifer [15] already pointed out the very close relationship between the robot morphology, the environment and the task. In fact, the morphology of the robot determines the capabilities it may display in a given environment: an optimized morphology for a given task simplifies control, makes the task easier and thus permits improving performance, whereas an inadequate morphology will lead to an inefficient solution to the task with the consequent increase in control complexity and computational cost.

Summarizing, there seems to be a close relationship between the physical design of a robot and that of its controller, as a function of the environment or environments and tasks to perform. This opens up a series of research questions of interest. The obvious one, which has been addressed by many authors such as [16]–[19], is how these relationships can be used in order to produce more efficient robots and controllers.

However, a probably more interesting and subtle question, that has generated less attention, is whether this morphology-control relationship or coupling can be used in developmental processes. That is, whether it makes sense to use it in order to improve the learning and adaptation capabilities of a robot that must address lifelong open-ended learning. In fact, it would be interesting to know whether this coupling provides any advantage when exploring complex behavioral spaces and, if so, under what circumstances.

In this paper we address this issue. In section II we provide some general ideas on morphological development. Section III is devoted to the presentation of the background on the application of morphological development to learning in quadruped and other types of walkers and we introduce the experimental setup we are going to use. The results of the experiments carried out are presented in section IV. Finally, we discuss these results in section VI and provide some conclusions and future lines of work in section VII.

## II. Morphological Development.

In humans, morphological development encompasses cognitive development [20], motor development [21] and body maturation [22]. Cognitive development consists in the creation of new world representations and in the consolidation and adaptation of those already present. On the other hand, motor development implies the continuous acquisition of motor skills to gradually improve the effectiveness of the movements. Finally, growth and maturation involve more than an increase in body size and weight. They imply other subtler physical aspects such as increases in muscle tone, bone mass, extensions in the motion range of the limbs and improved sensory capacity, among others.

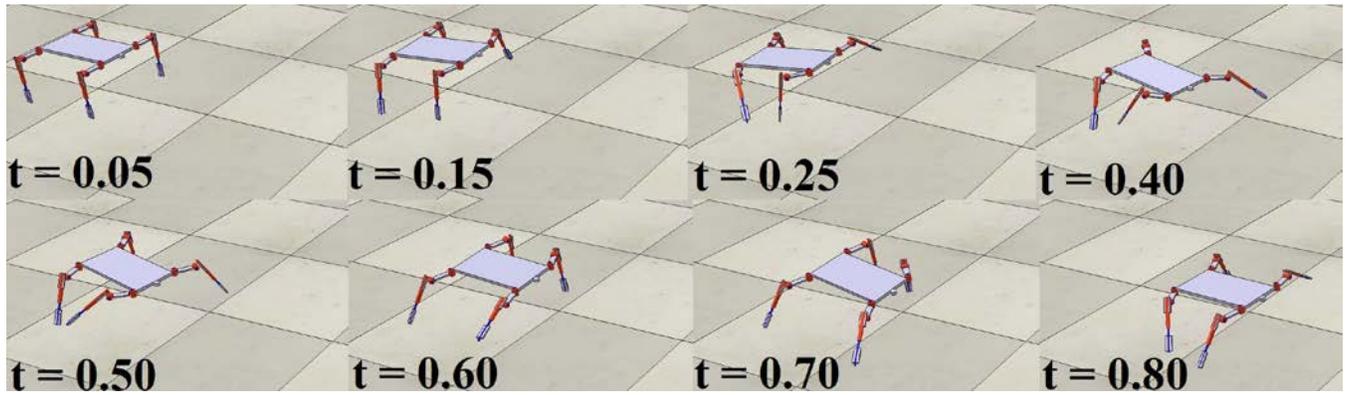

**Fig 1**. Snapshots of a motion sequence of the quadruped morphology used in the experiment with one of the controllers starting from a resting position. Each image has been taken at the specific simulation time (t) that is represented in seconds, in the lower left part of each frame. Each limb has three solid segments in a chain attached to the base by two revolute joints (red cylinders), which are actuated, and a linear joint (red rectangular cuboid), which is used for the morphological development.

Even though in psychology and biology these concepts have been addressed separately, different authors support that their interrelation and joint operation should be explored [23]. In fact, development is highly mediated not only by physical and cognitive changes within the body itself, but also by the influence that they have over the interaction with the environment and how this affects in the acquisition of new knowledge. Even though it is usually a continuous process, morphological development can be pictured as an evolution of the developmental process in a series of stages, from the simplest and starting stage, to the final and most complex one.

Looking at this from a mathematical and computational point of view, one way to facilitate searching for a solution in a complex search space is to initially consider a simplified version of the problem, as defined by the solution search space it defines [24], [25], and progressively add complexity as the algorithm is searching. In other words, we are producing a stage based approach in which each stage presents the algorithm with a solution space that is progressively more complex and more complete. What is important here is that in this type of stage based or curricular learning processes we also provide the algorithm with a starting point given by the best solution obtained in the previous stage, thus simplifying the incremental complexity of the subsequent task and deploying simplified and less computational costly solutions [26].

A division of the developmental process into stages also highlights an important difference between morphological development and cognitive development in the way how they handle the learning process. In cognitive development, the first units of knowledge to be learnt are basic features that will be used later as scaffolding to learn more complex knowledge nuggets that make use of them. This is not the case for morphological development. The cognitive structures created at each stage of morphological development are the starting point for learning the next stage. These structures are modified and adapted to the physical changes that occur during the morphological development process, giving rise to new cognitive systems that are better adapted to the new morphology (e.g. adult walking is not made up of different bits of baby walking and teenager walking). Thus the morphological development process steers a path through a series of solution spaces the system interacts with as it develops morphologically which lead to the final adult situation. The hypothesis here is that this path may simplify learning the final goal (infants have lower centers of mass, which facilitates learning to walk).

### III. EXPLORING THE APPLICATION OF MORPHOLOGICAL DEVELOPMENT.

After reviewing the main ideas underlying morphological development, we now want to focus on a particular robotic application in order to try to extract some conclusions on how morphological development can be applied. In particular, we are going to focus on a simple quadruped walking scenario over which we will carry out a series of experiments in order to test the effects of different parameters over learning using morphological development.

To the best of our knowledge, there are no previous references on strict morphological development in learning quadrupedal motion. We have found only two papers that addresses quadrupedal motion taking into account morphological development, one is from Bongard [27] and the other by Vujovic et al. [28], but, in both cases, in the framework of evolutionary robotics and evolutionary developmental robotics (EvoDevo). However, the work presented in these two studies is the one that best reflects our idea of development for learning. The aim of Bongard's study is to reach a source of light as fast as possible. The study was divided into 5 different experiments; each one consisted of 4 phases of evolution and each phase was associated to a specific state of the robots. He showed that employing development together with evolution in each phase contributes to faster success. Furthermore, evolving the morphology jointly with the behavior has demonstrated a more robust performance under environmental perturbations. However, it has also been shown that a combined approach based on morphological development and evolution can be detrimental and offer worse results than a case in which morphological development is not sought.

On the other hand, the objective of the study by Vujovic et al. is to compare the results obtained after applying an evolutionary or an EvoDevo sequence to a robot which needs to walk over flat terrain. In order to do that, the length and thickness of the legs are modified. They claim that the combination of evolution and development results in an improvement of the fitness in the experiment in some cases. However, this depends on the choice of developmental system. In fact, poorly chosen developmental parameters seem to result in poorer results than considering only evolution. The hypothesis the authors make is that reducing the search space in the wrong direction eventually removes the good solution choices that development could have made along the way.

We have found some papers, not many, on morphological development in the case of bipedal motion [29]–[31], and for hexapedal walking as we mentioned previously [27]. The results they present with respect to how morphological development influences learning are inconclusive. They provide examples where morphological development during the learning process improves over just learning with a fixed morphology, but in other examples the results are just the opposite.

In this work we are going to use quadrupedal walking as a first test case in an exploration of how to harness morphological development in an attempt to provide design guidelines to be able to use it appropriately. The main objective of this paper is to try to elucidate whether morphological development actually improves ANN learning and whether there is a difference when learning an ANN based controller in this scenario between morphological development through growth, which is closer to biological development, and morphological development through the liberation of degrees of freedom, which is the case most often contemplated in the literature. We also address the issue of whether the morphological development rate affects learning and, finally, we will provide some indications on the robustness achieved by the solutions obtained in each case.

## IV. EXPERIMENTAL SETUP.

The basic design of the quadruped that is going to be used in all the experiments is presented in Fig 1. It is made up of a central body of dimensions 30cm*15cm*1cm and 2kg, and 4 limbs, each one with two revolute joints and one prismatic joint. Each limb is composed of 3 segments, all of them present the same size and weight (5cm*2.5cm*0.5cm and 250g). The farthest segments of the robot's legs are joined by the prismatic joint. This joint has a maximum stroke of 7.5cm, which means that the length of the legs may vary from 10 to 17.5cm. The maximum rotation interval of the revolute joints is [-90, 90]. The controller of the robot is a neural network whose weights and structure are learnt using NEAT [32] specifically the MultiNEAT implementation [33]. It has one input and 8 outputs, each controlling the actuation of one joint. The input, is a sinusoidal function of amplitude 2.0 and frequency 1.0rad/s.

A series of learning experiments using NEAT have been run over implementations of the robot and environment using the VREP [34] simulator with the ODE [35] physical engine. Each NEAT learning run contemplates a population of 50 individuals and is trained for 300 generations. A total of 20 independent runs have been carried out for each experiment with the objective of gathering relevant statistical data. As the controller is obtained using NEAT, the learning strategy is based on a neuro-evolutionary process, where the fitness is the distance travelled by the robot. Each individual is tested for 3s with a simulation time step of 50ms and physics engine time step of 5ms.

In order to study the effects of development over a specific morphology we ran 3 different types of experiments:

*1) Reference experiment:* this experiment is run with a fixed morphology (the same as the final morphology for the rest of the experiments) from the beginning to the end. The robot starts at generation 0 with the maximum length of the legs and the neuro-evolutionary algorithm seeks a neural network based controller to achieve displacement.

*2) Leg growth experiment:* the robot morphology starts with the shorter version of the legs. That is, at the beginning the prismatic joint is fully contracted, its extension is 0, and the length of the legs is thus 10cm. The leg length is grown linearly with the generations until it reaches the maximum length of 17.5cms. This growth takes place in a set number of generations for each experiment. That is, the final morphology is reached at generation 20, 40, 60, 80, 100 and 120 depending on the experiment. This permits studying the relevance of the growth rate with regards to performance. In a way, we try to simulate the way knowledge is acquired by biological entities: their limbs grow until a certain age, and then learning continues with a fixed morphology.

*3) Motion range experiment (MR):* the robot starts with the final morphology, which is fixed during the whole experiment. What actually changes here is the motion range of each revolute joint, that is its maximum angle available. The angular development is applied on the denormalization of the output of the ANN, increasing gradually the maximum avaliable angle value.The ANN starts with a maximum available angle of half that of the final angle, and this maximum is increased linearly again, for a number of generations. In other words, in different runs the maximum possible motion range is reached at generation 20, 40, 60, 80, 100 and 120. This form of development is inspired on the DOF freeing mechanisms proposed in the literature [36]–[38], in which, at the beginning of development, the motion of the joints is locked or limited, progressively increasing their scope with time.

In addition to studying how morphological development affects the learning process, we have performed some experiments to try to determine whether there is any difference in the robustness of the results when faced with small changes in the morphology or motion range of the quadruped. With this objective, we have taken the best controllers of each type of experiment and tested their behavior in a series of quadrupeds with morphologies that were slightly different from those for which they were trained. These morphologies are based on the reference morphology of the quadruped, but two types of modifications were made:

1) *Growth modification:* The length of the legs was modified by 5% and 10% above and below the reference size.
2) *Angle modification:* the maximum motion ranges were modified by 5% and 10% above and below the reference angles.

## V. RESULTS.

The results of the training process for each case can be observed in Fig. 2 and Fig. 3. Fig. 2 is related to the experiments in which the legs are grown. The top graph displays the median of the best fitness obtained for the 20 independent runs at each generation for 3 of the growth rates and the reference. The shaded areas in the graph represent the areas between percentiles 75 and 25 for each experiment. Although we have carried out all of the experiments mentioned in section IV, for the sake of clarity in this graph, we only represent the experiments where the final morphology is reached at generations 20, 60 and 120 which are very representative of what is generally happening. The experiments that reach the final morphology at generation 20 obviously represent the fastest growth, whereas those that reach the final morphology at generation 120, the slowest one. In between, we included the experiment that reaches the final morphology at generation 60. It this particular case is also the one that produces the best results.

It can be easily observed that this morphological development mechanism based on leg growth provides better results than when there is no development. In fact, it can be clearly seen that all the developmental processes produce a noisy fitness curve while growth takes place. This is due to the fact that as the morphology is changing and the ANN that is being evolved needs to adapt to a changing body. But after growth ends, the curves become less noisy (the morphology is now fixed) and the system continues to adapt to its final morphology, improving its control over it and reaching fitness values that are in the order of 50% better than in the no-developmental reference case.

The statistical relevance of these results can be observed in the middle graph of Fig. 2. Each boxplot represents the median and the 75 and 25 quartile of each of the different types of experiments in the last generation. The whiskers are extended to 1.5 of the interquartile range (IQR). Single points represent values that are out of the IQR. All developmental samples are compared to the no-development sample. The statistical analysis has been carried out using the two-tailored Mann-Whitney test. We want to test whether the null hypothesis (if both compared samples are equal) is true. We consider a p-value of 0.05 as the significant value for accepting or rejecting the null hypothesis. Asterisks and lines refer to the statistical difference between the samples represented and its change from numerical values to a graphic representation for simplicity can be seen in the legend of the figure itself. The most relevant result is obtained for growth until generation 60, obtaining a p-value 0.00195, which means a huge rejection of the null hypothesis. That is, in this case morphological development clearly improves learning with respect to no-development. Growth experiments up to generations 80 and 100 present p-values of 0.02227 and 0.01058 respectively, showing that both cases of development offer statistically relevant better results than no development. In the case of growth up to generation 20 (p-value: 0.08103), generation 40 (p-value: 0.07205) and generation 120 (p-value: 0.06787) it is considered that the null hypothesis cannot be rejected because their p-values surpass the reference p-value, even though they are very close to this limit. Consequently, they are not statistically different from no development. All of these results provide a first indication that the growing rate needs to be aligned with the learning time in order to be able to get better results over no development. This is probably because the system is generally changing too fast for the ANN controller to be able to follow it. On the other hand, growing too slowly probably allows the ANN to fall into deep local minima from which it is difficult to exit as the morphology changes.

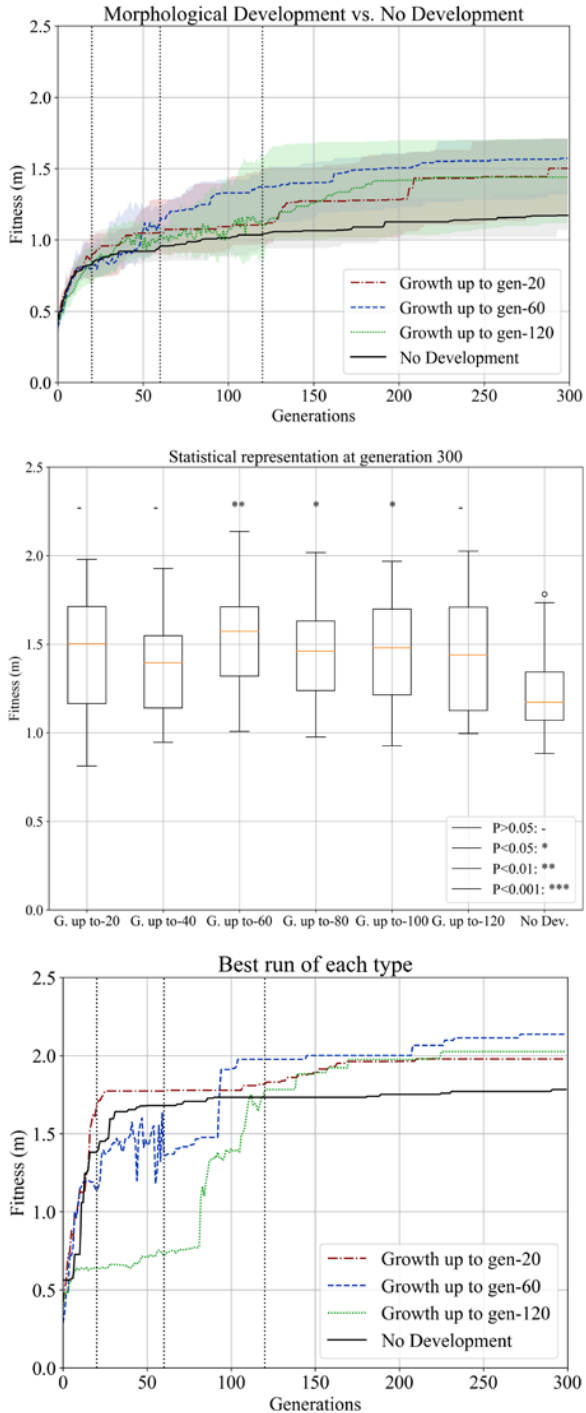

**Fig 2.** Top: Statistical representation of the neuroevolution of the 20 independent experiments for the growth of the legs until generation 20, 60, 120 and for the case of no-development. Middle: Statistical representation of the performance obtained from the 20 independent experiments at the end of the neuroevolutionary process. Bottom: Behavior of the best individual for each type of experiment.

Fig. 2 bottom represents the evolution of the best individual for each kind of experiment. We can observe how the curve that shows the best learning process in the case of no development quickly stagnates in fitness from generation 50 onwards. This shows that, from that generation it is hard to find individuals that improve the existing ones and thus, improving performance is also difficult. A similar behavior can be observed in the best individual of the experiment that grows until generation 20. It shows a stagnation during most of the evolutionary process except in the interval approximately between generations 100 and 170, where there is improvement in the performance of the controller. Development graph up to 120 also shows stagnation at the end, but shows more progressive learning both during the growth stage and with the final morphology onwards. Remarkable by the endo of the growing phase, the fitness value in all the cases has not significantly improved on the no development case. It is only after the morphology has stabilized that the relevant fitness improvements take place, hinting at the importance of the growth stage in order to avoid local minima and position the system in areas of the solution space that allow for improvement. This is, in fact, the case of growth until generation 60 for the best individual. The graph shows an erratic fitness improvement until reaching the final morphology, from where it stabilizes and begins progressive more stable learning process, being the fitness obtained at the end of the learning process (2.14m) much higher than the one obtained in the case of no growth (1.78m).

Fig. 3 corresponds to the experiments that use angular motion range variation as morphological development. As can be seen, although the method of liberating of degrees of freedom is the morphological development method most often used in the studies found in the robotic literature (see [38], [39]), in this case there was no statistically relevant improvement in the experiments that used this type of morphological development when compared to the reference case. Although, there are isolated individuals which appear to offer better performance than individuals in the case of no-development, they general results are statistically similar. Development up to generation 60 results in the best performance, with a p-value of 0.42488 which is far from the 0.05 necessary to be considered a significant difference.

In order to analyze the robustness of the selected approach to morphological development, the 20 best independent individuals obtained from each experiment were tested over the morphological variations indicated in section IV. Other authors have also studied the concept of robustness, although, it is not clear how the concept of robustness is defined. It is usually addressed as the performance under a novel environment, however this is a vague definition

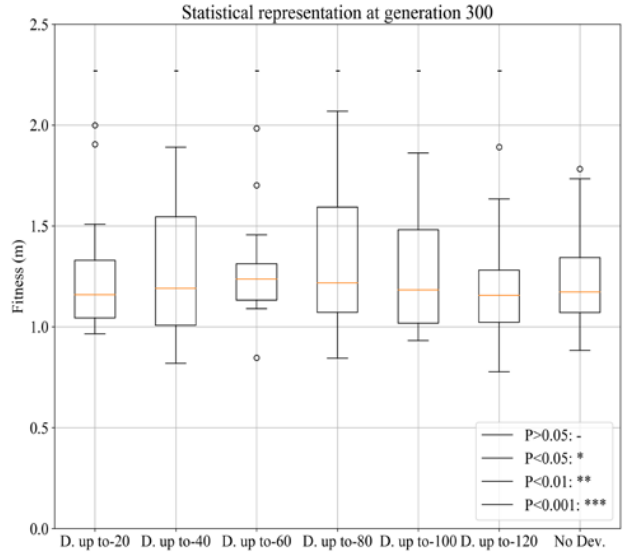

**Fig 3**. Statistical representation of the performance obtained from the 20 independent experiments at the end of the 300 generations considering morphological development as increasing the that makes it difficult to compare the term robustness between experiments. In [27], where the task objective consisted in reaching a source of light as fast as possible, robustness is defined as the relative performance of the controller under slight random perturbations during its movement. In [40] the task objective consists in grasping objects of different sizes. Robustness, in this case, is defined as the relative performance of the controller versus grasping objects that present small differences in dimensions compared to the objects considered during learning.

Thus, given that there is no clear universally accepted definition of robustness, in our case, we have applied a definition related to the processes involved. As a consequence, we have defined robustness as the relative performance of the controller under small variations in the dimensions of the legs and variations in the maximum motion range that a joint can reach. The selected controllers for testing robustness are the ones that offered the best results for each kind of morphological development process (growth up to generation 60 and angular development up to generation 60) and the no-development case as a reference.

Fig. 4 displays the results of these experiments. To produce these results, the best controller obtained for each case in each of the runs was tested over the morphological and motion range variations. The fitness of each individual was normalized using the fitness achieved during the learning stage. A statistical representation of these relative fitness values for 20 runs in each of the three cases (NoDev, Growth up to 60 generations and motion range up to 60 generations) for the two morphological variations are presented. The normalization explains why the central case in each case presents a constant value of 1. The top graph represents the statistical results considering a 5% and 10% increment and decrement in the percentage of the maximum range of joint angle available. The reference performance obtained in the neuroevolutionary process is shown in the middle of each group and described with a dash in the label, indicating that there are not modifications to the final morphology the controller was adapted to. Fig. 4 bottom displays the statistical results considering a 5% and 10% increment and decrement in the leg length with respect to the final length. Visually analyzing the results presented in Fig. 4, in the case of morphological variations that involve both an increase in the extension of the legs and modifications in the motion range available, we cannot say that one case of morphological development is more robust than another or better than the no-development case for the definition of robustness we have chosen. In fact, practically all the medians, are within the range from 1 (reference value) to 0.8 of the normalized value, presenting the obvious behavior whereby the larger the morphological difference with respect to the reference morphology, the worse the results obtained.

## VI. DISCUSSION.

We hypothesize that the learning improvement obtained with morphological development is due to three main factors:

1) By reducing the space of possible solutions that allow the robot to move, it is easier to find solutions that allow it to move. This does not mean that the solutions found must necessarily be valid in the final morphology, but they are solutions for the morphological situation of the moment that allow the development of a stepped and progressive learning sequence.

2) In this particular case, considering the quadruped, we believe that morphological development offers a second advantage for the morphology, and that is that the initial configurations offer greater stability to the quadruped at the beginning of learning, allowing it to remain on its 4 legs without falling. This favours a larger number of valid solutions at the beginning and as growth occurs, those solutions that best adapt to the learning process are selected.

3) It increases exploratory behavior. As can be seen in the best execution for the case of no-development in Fig. 2 bottom, the evolution converges rapidly to a suboptimal solution, becoming quickly stagnated. This shows the existence of local optima in the solution space of the problem that are far from being the best solution that can be achieved. The best executions obtained considering morphological development up to 60 and 120 generation show a radically different behavior. Both show an erratic behavior at the beginning, with ups and downs in the value of fitness, typical of exploratory behavior, but once the final morphology is reached, the behavior improves progressively and favourably, until in surpasses the maximum no-development fitness. This highlights the benefits of this initial exploration as compared to the rapid convergence in the case of no-development or even in the cases of very fast growth (the cases where growth takes place up to 20 or 40 generations).

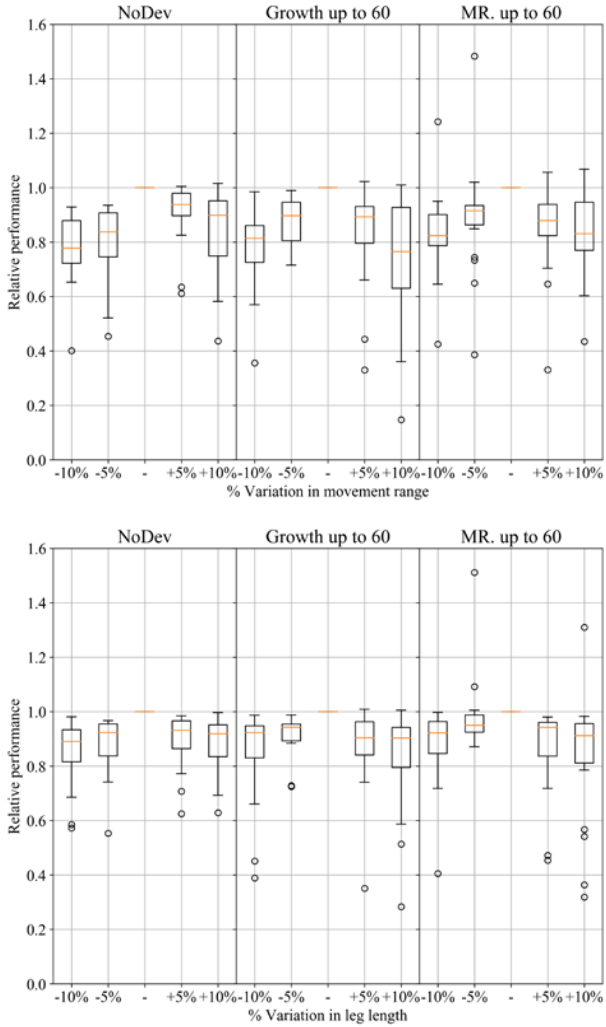

**Fig 4.** Robustness analysis. Top: relative performance considering a proportional variation of the maximum motion range available. Bottom: relative performance considering a proportional variation of the leg length.

On the other hand, as shown in Fig. 3, morphological development through the release of angle constraints has not been shown to offer better results than the case of no-development. Indicating that not any type of morphological development is valid for a given morphology. Something that is also observed in development through growth, as no relevant improvement over the no development case can be observed for growth rates up to generation 40.

These results show the need to study in greater depth the underlying mechanisms in morphological development applied to robotics, in order to find out in which cases it favours learning or in which it is irrelevant, as we can see in Fig. 3, or even harmful.

VII. CONCLUSIONS.

In this paper, we have shown that ontogenetic morphological development while a robotic quadruped is learning a locomotion task may help to find better solutions when compared to a robot that does not follow any kind of development. Specifically, we have found that the morphological growth of the limbs helps to learn better controllers if the growth rate is chosen correctly. On the other hand, while the angular development of the joints has not improved the learning, it did not present a negative effect.

Regarding the robustness of controllers when handling morphological perturbations, the morphological development shows an equivalent reduction in performance to that of the no development experiment. However, due the fact that the morphological development controllers present higher performance, the controllers obtained by morphological development still behave better in the cases for which they were not designed that than those where no development was used.

Summarizing, from these results it seems that certain types of morphological development within certain development rates can really help in learning better controllers for complex tasks. Nonetheless, much more work is needed over many other tasks to really be able to provide effective engineering indications on how to apply morphological development. We are currently extending the range of cases and complexity in order to study this issue.

ACKNOWLEDGMENT.

This work has been partially funded by the Ministerio de Ciencia, Innovación y Universidades of Spain/FEDER (grant RTI2018-101114-B-I00), Xunta de Galicia and FEDER (grant ED431C 2017/12) and M. Naya-Varela is very grateful for the support of the

UDC-Inditex 2019 grant for international mobility. We also want to thank CESGA (Centro de Supercomputación de Galicia. https://www.cesga.es/) for the possibility of using its resources.